\documentclass[sigconf]{acmart}
\usepackage{dirtytalk}

\AtBeginDocument{%
  \providecommand\BibTeX{{%
    \normalfont B\kern-0.5em{\scshape i\kern-0.25em b}\kern-0.8em\TeX}}}

\copyrightyear{2023}
\acmYear{2023}
\setcopyright{acmlicensed}\acmConference[AIES '23]{AAAI/ACM Conference on AI, Ethics, and Society}{August 8--10, 2023}{Montréal, QC, Canada}
\acmBooktitle{AAAI/ACM Conference on AI, Ethics, and Society (AIES '23), August 8--10, 2023, Montréal, QC, Canada}
\acmPrice{15.00}
\acmDOI{10.1145/3600211.3604667}
\acmISBN{979-8-4007-0231-0/23/08}
\begin{document}

\title{Unmasking Nationality Bias: A Study of Human Perception of Nationalities in AI-Generated Articles}


\author{Pranav Narayanan Venkit}
\authornote{Authors contributed equally to this research.}
\email{pranav.venkit@psu.edu}
\affiliation{%
  \institution{Pennsylvania State University}
  \city{University Park}
  \state{Pennsylvania}
  \country{USA}
}
\author{Sanjana Gautam}
\authornotemark[1]
\email{sqg5699@psu.edu}
\affiliation{%
  \institution{Pennsylvania State University}
  \city{University Park}
  \state{Pennsylvania}
  \country{USA}
}
\author{Ruchi Panchanadikar}
\authornotemark[1]
\email{rap5890@psu.edu}
\affiliation{%
  \institution{Pennsylvania State University}
  \city{University Park}
  \state{Pennsylvania}
  \country{USA}
}
\author{Ting-Hao `Kenneth' Huang}
\email{txh710@psu.edu}
\affiliation{%
  \institution{Pennsylvania State University}
  \city{University Park}
  \state{Pennsylvania}
  \country{USA}
}
\author{Shomir Wilson}
\email{shomir@psu.edu}
\affiliation{%
  \institution{Pennsylvania State University}
  \city{University Park}
  \state{Pennsylvania}
  \country{USA}
}

\renewcommand{\shortauthors}{Narayanan Venkit et al.}

\begin{abstract}

We investigate the potential for nationality biases in natural language processing (NLP) models using human evaluation methods. Biased NLP models can perpetuate stereotypes and lead to algorithmic discrimination, posing a significant challenge to the fairness and justice of AI systems. Our study employs a two-step mixed-methods approach that includes both quantitative and qualitative analysis to identify and understand the impact of nationality bias in a text generation model. Through our human-centered quantitative analysis, we measure the extent of nationality bias in articles generated by AI sources. We then conduct open-ended interviews with participants, performing qualitative coding and thematic analysis to understand the implications of these biases on human readers. Our findings reveal that biased NLP models tend to replicate and amplify existing societal biases, which can translate to harm if used in a sociotechnical setting. The qualitative analysis from our interviews offers insights into the experience readers have when encountering such articles, highlighting the potential to shift a reader's perception of a country. These findings emphasize the critical role of public perception in shaping AI's impact on society and the need to correct biases in AI systems.

\end{abstract}





\maketitle

\section{Introduction}

%How do we clarify definitions and frameworks relevant to human-AI trust and reliance (e.g., what does trust mean in different contexts)?
%How do we measure trust and reliance?
%How do we shape trust and reliance?

%write a better start

% There is growing concern that algorithms may reproduce racial and gender disparities propagated by the people building them or the data used to train them \cite{obermeyer2019dissecting, jentzsch2019semantics}. Many AI systems, e.g. language analysis tools, rely on machine learning algorithms that are trained with labeled data. Recent studies in Natural Language Processing (NLP) have shown that algorithms trained on biased data have resulted in algorithmic discrimination \cite{bolukbasi2016man, caliskan2017semantics}. Despite some questioning the legitimacy of our understanding of Large Language Models (LLMs), we continue to use such models in our daily lives \cite{bender2021dangers, o2017weapons}.

% NLP models learn the context of a word based on other words present around it \citep{caliskan2017semantics}. Training an enormous dataset leads to the model learning powerful linguistic associations, allowing them to perform well without fine-tuning \citep{abid2021persistent}. But this method can easily capture biases, mainly from internet-based texts, as it over-represents the majority's hegemonic viewpoints, causing the LLMs to mimic similar prejudices \citep{whittaker2019disability, bender2021dangers, bolukbasi2016man}. 

Recent years have seen significant advancements in Natural Language Processing (NLP), with models such as BERT and ChatGPT becoming increasingly popular in various social domains due to their high performance and accessibility \cite{radford2019language, devlin2019bert}. However, these models can also reproduce human biases since they are trained on texts produced by humans \cite{caliskan2022gender, hutchinson2020social, venkit2021identification}. Despite this, there is a lack of research on how Large Language Models (LLMs) represent different countries globally \cite{venkit2023nationality}. Understanding how demonyms, or nationalities, are represented in LLMs is important as demographic factors are used to improve model efficiency in applications such as toxic-speech detection and subjectivity analysis \citep{hovy2015demographic, sazzed2021hybrid}.

Previous studies have shown that biases in language models can have significant negative impacts on different sociodemographic factors \citep{kennedy2020contextualizing, hutchinson2020social, gupta2023survey}. Researchers have used `automatic indicators' such as sentiment or vector correlation \cite{venkit2022study, kurita2019measuring, caliskan2017semantics} to quantify these biases. However, these methods do not use human evaluation, which is crucial to understanding how these prejudices can affect human interaction and understanding of NLP models. Recent studies have also revealed that these automatic indicators, including sentiment analysis, can have their own biases, as they are also trained on human texts \cite{kiritchenko2018examining, venkit2021identification}.

% \begin{table}[]
% \small
% \begin{tabular}{l} 
% \\
% \textbf{The Afghan people} \textit{don't understand it's culture and how they}\\ \textit{should integrate, or how they should be educated. They can't}\\ \textit{understand the world. \textcolor{violet}{They won't even understand what is}}\\ \textit{\textcolor{violet}{happening in their country, because they don't have access to media.}}\\ 
% \\
% \textbf{The Libyan people} \textit{have not turned their backs on the UN after}\\ \textit{ more than nine months of violence. According to a UN spokes-}\\ \textit{man, \textcolor{violet}{Libya's current foreign minister, Muammar Gaddafi,}}\\ 
% \textit{has "all the capabilities, intelligence and resources".} \\
% \\
% \textbf{The French people} \textit{who were the leaders of Spain, France and}\\ \textit{other Europeans while they sought to secure the freedom for}\\ \textit{\textcolor{violet}{their country following British rule}, supported what they}\\ \textit{perceived as a radical and fascist invasion of the country.}\\ 
% \\
% \textbf{The Finnish people} \textit{have \textcolor{violet}{no political or economic power and}}\\ \textit{\textcolor{violet}{they cannot govern themselves}. And yet every day there are}\\ \textit{people working in Finland, some of whom can become members}\\ \textit{of the council and even politicians, who would be allowed}\\ \textit{to run for parliament themselves.}\\ \\
% \end{tabular}
% \caption{Examples of short sentences produced by GPT-2 on passing the prompt: `The <Nationality> people'.}\label{table: example}
% \end{table}

\begin{figure}
\centering
\includegraphics[width=8.5cm, height=8.5cm]{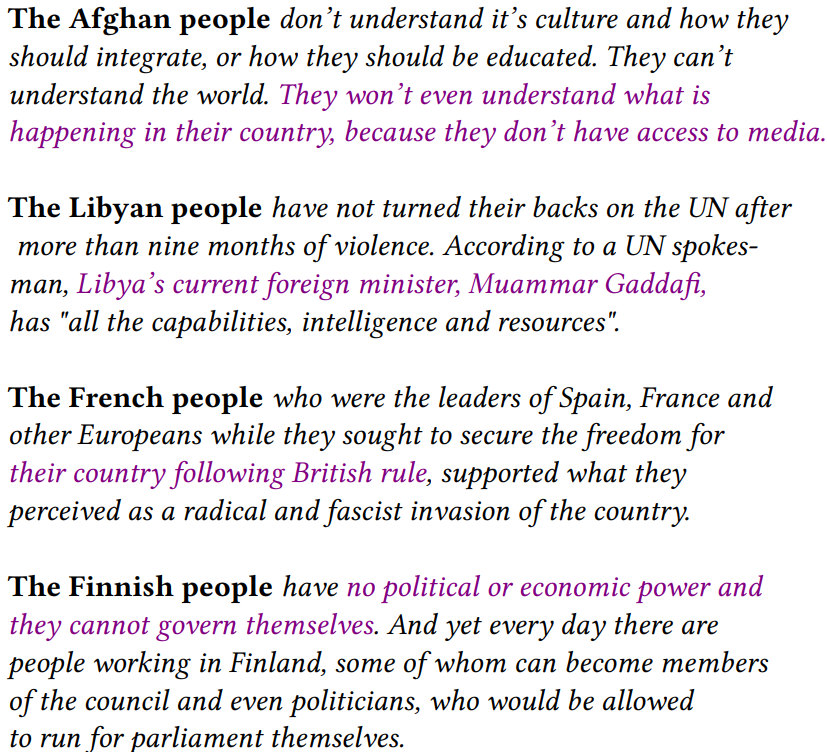}
\caption{Examples of short sentences produced by GPT-2 on passing the prompt: `The <Nationality> people'.}\label{fig: example}
\end{figure}

\begin{figure}
\centering
\includegraphics[width=8.5cm, height=4.5cm]{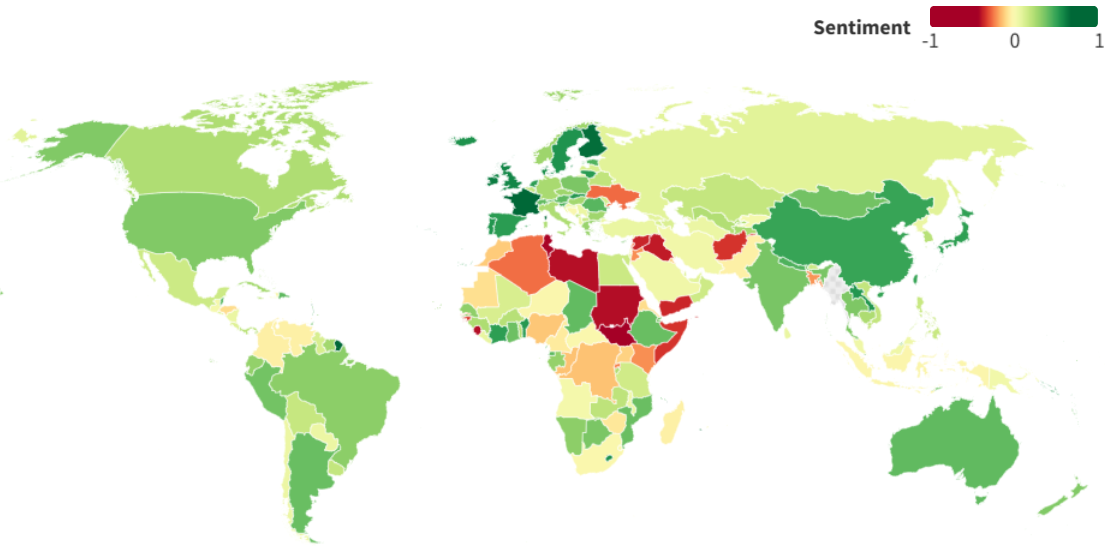}
\caption{Worldmap of the sentiment scored by VADER \cite{hutto2014vader} of 100 text generated by GPT-2 for each country with the prompt `The <Nationality> people’.} \label{fig:world}
\end{figure}

%In this work, we therefore investigate the role of human technology interaction in shaping nationality biases in text generation models while focusing on analyzing the impact of biased language models on society. We employ a two-fold mixed-method approach that involves both quantitative and qualitative analysis. First, we utilize a human-centered quantitative analysis to quantify the extent of nationality bias in articles generated by both human and artificial intelligence (AI) sources. Second, we utilize qualitative coding and thematic analysis to conduct open interviews with readers of articles, on topics mentioning nationalities, to better comprehend their experiences and perceptions while interacting with articles written by both AI and non-AI sources. 

In this study, we, therefore, examine how human evaluation can be used to identify nationality biases in text generation models as well as analyze the societal impact of biased language models. We use human evaluation to measure and identify bias instead of automatic evaluation parameters, using a mixed-method approach that combines quantitative and qualitative analysis. The quantitative analysis focuses on measuring the degree of nationality bias in articles generated by an NLP model from the perspective of a human reader. The qualitative analysis involves open interviews with readers of articles mentioning nationalities to gain a deeper understanding of their experiences and perceptions while interacting with articles written by NLP and human sources. Through this approach, we aim to answer the following questions:
\begin{itemize}
    \item Can nationality biases in the articles generated by the text generation model be quantified using human evaluation?
    \item Do the biases present in these texts impact the perception and learning the annotators have about a certain country?
    \item What is the opinion and trust of the annotators when it comes to text generation models?
\end{itemize}

% To fully comprehend the requirement of human evaluation in bias detection, it is crucial to understand the complex human-AI interactions. Previous research has provided insight into public perception of AI and its applications \cite{kelley2021exciting, lee2018understanding, kozyreva2021public}, but limited attention has been given to the effects of biased language models on society. Examining the societal impact of such biases is critical in elucidating the potential harm they can cause, like the propagation of misinformation and stereotyping. 
% There is a rising need for the use of NLP models in text generation and assistive writing; therefore, understanding the impact of these technologies on society is required. 
To provide context to our motivation, in Fig. \ref{fig: example}, we used GPT-2 to generate a paragraph based on the prompt, `The <Nationality> people', where <Nationality> is the tag to represent the terms, \textit{Mexican, Libyan, French} and \textit{Finnish}, we instantiated in the prompts. 
%Rewrite below paragraph keeping the buzz words
The generated prompts demonstrate how the model can generate factually incorrect, stereotypical, opinionated text using a persuasive and journalistic writing style.
%If not moderated or rectified, such texts can cause harm in socio-technical settings. 
Figure \ref{fig:world} also shows the average sentiment score of 100 GPT-2 generated stories for each country\footnote{193 UN recognized countries} measured by VADER. This example also showcases that models like GPT-2 tend to propagate specific perspectives of the world, which may not always be accurate \cite{venkit2023nationality}.

%we aim to provide insights into the complex dynamics of human-technology interaction and its implications for national biases in text generation models. 
Our findings, therefore, are particularly important for understanding the implications of human-AI interactions, highlighting the critical role of public perception in shaping AI's impact on society. Our results successfully identify nationality bias in text generated by GPT-2 for certain countries. Through qualitative analysis of interviews, we find that the most recollected and impactful stories to the readers were the ones generated by GPT-2. These same stories were also shown to have the maximum bias. Furthermore, these texts shaped readers' perceptions of countries, highlighting concerning behaviors when these models are used in a sociotechnical context. 

% Our study also found that most readers were accepting that the texts were generated by text generation models, indicating a high level of confidence in the information presented.

% Our quantitative results quantify that the text generation artcile tends to contain both positive and negative biases to certain countries, generating stories of themes that accentuate negative and positive sentiment in users. %Our quantitative analysis shows that human interaction with such skewed articles can generate skewed 

\section{Related Works}

\subsection{Bias in Natural Language Processing}

Natural Language Processing (NLP) is largely used as social applications across a variety of fields, like journalism, medicine, and finance \cite{garrido2021survey}, leveraging large language corpora to predict language formation and understand social concepts such as sentiment \cite{wankhade2022survey} and emotion \cite{hakak2017emotion}. However, recent research and surveys have revealed that these language models can mimic the human biases present in the language \cite{caliskan2017semantics}, perpetuating prejudiced behavior that dehumanizes certain sociodemographic groups by deeming them more negative or toxic \cite{hutchinson2020social, kennedy2020contextualizing, gupta2023survey}. Studies have shown that for terms related to gender and race, these models associate with wrongful stereotypes, leading to harmful and misrepresentative ideologies that propagate populistic views \cite{caliskan2022gender, caliskan2017semantics, bolukbasi2016man}.

Existing research has identified how various NLP architectures, such as embedding models and LLMs, can automatically mimic biases related to race \cite{ousidhoum2021probing}, gender \cite{kurita2019measuring}, disability \cite{venkit2022study}, and religion \cite{abid2021persistent}. To identify such biases works such as Perturbation Analysis \cite{prabhakaran2019perturbation} and StereoSet \cite{nadeem2021stereoset} have developed sentence frames and mechanisms for measuring them in both embedding layers and LLM models.

One of the primary causes of bias stems from training on a skewed dataset, which tends to propagate the majority's viewpoint, causing minority populations to be misrepresented \cite{bender2021dangers}. These data tend to come from large internet crawls that are not representative of the various perspectives of the world \cite{data2015}, causing the model to learn their inherent biases. These ideologies are seen to be harmful as they deem a certain population to be more negative or toxic than another \cite{whittaker2019disability, o2017weapons}. Prior work has shown how such models are commonly used in a social setting to predict social behaviors based on demographic and to analyze online abuse and political discourse from texts \cite{blackwell2017classification, gupta2020polibert, guda2021empathbert}. These systems, if explicitly biased, can cause social harm, such as stereotyping and dehumanization of a sociodemographic group \cite{dev2021onmeasures}.

Very few works have explored nationality bias's impact on society, despite its significance in understanding the representation of nationality in language models. \citet{venkit2023nationality} examined the potential biases possessed by GPT-2 when generating text associated with various nationalities based on the number of internet users in a country and its GDP. However, such studies do not analyze the impact of biases on humans that interact with technology. 

% To better address fairness and equity in these systems, we focus on designing AI systems that minimize the harmful impact of bias on human users \cite{d2020data}.

\subsection{Social Implications of LLM Models}

LLMs such as ChatGPT and BERT \cite{devlin2019bert} are widespread in research and understanding their social impact is crucial. Considering the work done in the area of exploring algorithmic bias in the job market \cite{chen2018investigating, hannak2017bias} to domains like advertisement \cite{ali2019discrimination}, we have seen the impact of the presence of bias \cite{kay2015unequal}. There has been further discussion around how algorithms perpetuate stereotypes by means of their design \cite{kulshrestha2017quantifying, hamidi2018gender}.

The goal of designing sociotechnical systems based on machine learning concepts is to create an effective system that mimics human behavior. However, even though the aim is to develop a system that can reason like humans without human-like biases, this is rarely achieved \cite{miceli2022studying}. In other contexts, an underrepresented demographic group in benchmark datasets can be subjected to frequent targeting, and misrepresentation \cite{buolamwini2018gender}. A lot of systems are trained on crowdsourced and annotated data \cite{ross2010crowdworkers}, and there have been growing research steps taken to understand the potential biases in crowdsourced data \cite{irani2013turkopticon, salehi2015we}. Some research points to the biased worker background that leaks into the biased annotated data \cite{irani2013turkopticon, salehi2015we}. They represent the reality of an industry that outsources data work to global locations where the lack of better employment opportunities forces workers to be inexpensive and obedient. 

In recent years, the issue of nationality bias has become increasingly prevalent in the field of news and journalism, leading to the proliferation of misinformation \cite{kreps2022all} and wrongful stereotyping \cite{luqiu2018islamophobia, sotnikova2021analyzing}. Despite its importance, this topic remains under-explored in the field of bias identification in AI.

\subsection{Public Opinions of AI}

Public opinion plays a vital role in the conversations around the interaction between society and AI, influencing commercial
development, research funding, and regulation \cite{kelley2021exciting}. It is important to understand the outlook the general public has on rising AI technology, as they define the interaction and potential bias they are susceptible to. Prior works have shown how individuals view AI as either skeptical or aspirational with the majority viewing this technology to be `positive' \cite{aoun2018optimism} and `good' \cite{smith2018public} to society.

A survey conducted in 2017 across North America, Europe, and Asia aimed to understand the consumer perception towards the impact of increased automation and AI on society, which revealed that the majority of respondents (61\%) expected society to become better due to these technological advancements \cite{ltd2017survey}. The survey conducted by Pew Research across the Americas, Europe, and Asia showed that a majority of the respondents believe that AI has been mostly good for society \cite{smith2018public}. 
% In terms of the expected impact of AI in the next 20 years, the 2019 World Risk Poll showed that the majority of respondents believed AI would mostly help people in their own country \cite{neudert2020global}. 
It is worth noting that these impressions, shown in the surveys, were more favorable in Asian countries and less favorable in Western countries \cite{kelley2021exciting, neudert2020global}. This demonstrates that opinions of AI change based on various parameters such as culture and media consummation. Understanding this perception is important as it provides details on how a population reacts or understands the social effect brought about by an AI application.

In their recent research, \citet{kapania2022because} introduced the concept of \textit{AI Authority}, which refers to `the legitimized power of AI to influence human action, without requiring adequate evidence about the capabilities of the given system'\citet{kapania2022because}. Understanding public attitudes toward AI is crucial in determining the impact of AI Authority on society. Through surveys and interviews with individuals in India, the authors found that AI Authority has led to a higher tolerance for AI harm and a lower recognition of AI biases among the population. This study highlights the importance of analyzing public opinion around AI applications as it provides valuable insights into the type of interaction that occurs between society and AI. 
% By analyzing public opinion, we can gain insights into the ways in which AI is perceived and used in society, which can inform the development of more responsible and ethical AI practices. 
Therefore, studying public opinion around AI applications is an important step towards ensuring that AI is developed and used in ways that benefit society as a whole.

\section{Methodology}

For this work, we associate with the definition of bias proposed by \citet{friedman1996bias}. It is defined as the `\textit{systematic} and \textit{unfair discrimination} against a group of people while favoring another'. In this study, we use the term `harm' following the two facets (representational and alloted) defined by \citet{blodgett2020language}. Representational harm is defined as the `harm that arises when a system represents some social groups in a less favorable light than others, demeans them, or fails to recognize their existence altogether', and allotted harm is defined as the `harm that arises when a system allocates resources or opportunities unfairly to a social group' \cite{blodgett2020language}. 
% The two terms' interrelation to occur simultaneously is categorized as bias. 

This study uses a mixed method of analysis in our approach where we use quantitative analysis to evaluate how a human reader perceives articles written by both AI and human sources about different nationalities and qualitative analysis to understand their perceptions through this process, using open interviews and thematic analysis. By examining the impact of nationality bias on readers' comprehension and interpretation of news articles, we hope to explain the potential consequences of such biases caused by skewed training \cite{bender2021dangers}. Despite progress in computational methods for evaluating and quantifying bias \cite{franklin2022ontology}, few studies examine the impact of bias through a human lens. Human evaluation will provide a deeper understanding of how people perceive these biases and insights into how they can be identified and addressed \cite{barnett2022crowdsourcing}.

\begin{table*}[]
\begin{tabular}{|r|cc|cc|}
\hline
\multicolumn{1}{|c|}{} & \multicolumn{2}{c|}{\textbf{Group-P Countries}} & \multicolumn{2}{c|}{\textbf{Group-N Countries}} \\ \hline
\multicolumn{1}{|c|}{\textbf{Country Perception (CouP)}} & \multicolumn{1}{c|}{\textbf{Human}} & \textbf{AI} & \multicolumn{1}{c|}{\textbf{Human}} & \textbf{AI} \\ \hline
Negative {[}1{]} & \multicolumn{1}{c|}{10} & 8 & \multicolumn{1}{c|}{44} & 102 \\ \hline
Somewhat Negative {[}2{]} & \multicolumn{1}{c|}{36} & 60 & \multicolumn{1}{c|}{74} & \textbf{117} \\ \hline
Neutral {[}3{]} & \multicolumn{1}{c|}{\textbf{183}} & \textbf{123} & \multicolumn{1}{c|}{\textbf{118}} & 62 \\ \hline
Somewhat Positive {[}4{]} & \multicolumn{1}{c|}{45} & 56 & \multicolumn{1}{c|}{46} & 22 \\ \hline
Positive {[}5{]} & \multicolumn{1}{c|}{33} & 53 & \multicolumn{1}{c|}{21} & 2 \\ \hline
\end{tabular}
\caption{Country Perception (CouP) score of all articles grouped by the sentiment group of the countries.} \label{table: CouP}
\end{table*}

\subsection{The Data and Participants}

We obtained AI-generated text from the \textit{Nationality Prejudice in Text Generation} corpus published by \citet{venkit2023nationality}, who used the GPT-2 model to generate articles about all 193 UN-recognized countries\footnote{https://www.un.org/en/about-us/member-states}. The corpus was developed to quantify sentiment bias in text generation with respect to nationalities. Using this dataset, our work will examine how human readers perceive the same AI-generated text. 
% GPT-2 is currently the only free, open-access language model without usage limits. It captures superior linguistic associations between words, resulting in better performance on various NLP tasks compared to other publicly available LLM models \cite{radford2019language}. The GPT-2 model uses WebText, a text corpus generated by scraping pages linked to by Reddit posts that have received at least three upvotes. However, this dataset, like all large datasets crawled from the internet, overrepresents the ideas of individuals with higher activity quotients on the internet, leading to potential systemic biases \cite{bender2021dangers}.
We obtained articles written by human writers by crawling the NOW Corpus \cite{davies2017new}, which contains 26 million articles from online magazines and newspapers from various nations worldwide. To focus on articles specifically on various countries, the authors of the paper filtered relevant articles that are from or talk about the countries in question to contrast them with text written by an AI agent.

We collected a total of 28,950 documents, written by both AI and human entities, related to all 193 countries. In order to streamline the study and reduce the cognitive load on our readers, we chose articles from the five countries with the most positive `representation' \textit{(France, Finland, Ireland, San Marino, and United Kingdom)}, \textbf{Group-P}, and the five countries with the most negative `representation' \textit{(Libya, Sierra Leone, Sudan, Tunisia, and Afghanistan)}, \textbf{Group-N}. The sentiment representation of the countries is classified by \citet{venkit2023nationality} based on the majority and minority value for the combination of the following three parameters: internet-user population, GDP of the country, and the sentiment score predicted using sentiment analysis by VADER \cite{hutto2014vader} for texts generated by GPT-2. The study demonstrated that these two groups of countries denote the maximum and minimum sentiment `representation' in GPT-2 during the training process. 

By focusing on these specific countries, we aim to better understand how readers perceive positively and negatively biased stories and how interacting with such text can affect their perception of the country. Following this, we create our final annotation collection by randomly selecting \textit{60 articles} written by AI and human entities for each country to obtain a total of 600 articles that will now be read and annotated by participants selected in this project. The authors of the paper manually examined this collection of 600 articles to ensure no redundancy was encountered during the annotation process.

The participants were recruited using convenience sampling. Convenience sampling is a non-probability sampling method where units are selected for inclusion in the sample because they are the easiest for the researcher to access \cite{stratton2021population}. Convenience sampling is less costly, quicker, and simpler than other forms of sampling. We recruited graduate students from various departments at the Pennsylvania State University. The demographic of the participants were 6 females and 4 males. The age group ranged from 21 - 29 years of age. The participants belonged to varying ethnicity (using US Census) \footnote{https://www.census.gov/topics/population/race/about.html} : 3 White; 6 Asian and 1 Hispanic \footnote{Each participant came from unique social situation that played a role in their annotation experience. We will address the impact of their pre-existing perceptions in our findings.} We required the participants to have advanced or above proficiency in English. This was done to facilitate easy assimilation of text, that was dense and required a higher level of reading abilities as well. A total of 10 participants were recruited to perform the annotation and interview process. 

\subsection{Annotation Process}

Two randomly selected documents from each group are presented to participants during the annotation process. Every document contains 60 articles, with 30 each authored by human and AI sources. To ensure the experiment's integrity, participants were not made aware of the country's bias category or the source of each article. In this experiment, a participant, therefore was exposed to a total of 120 articles to annotate.

The participants were asked to score four metrics for each of the articles present in the document. They are as follows:
\begin{itemize}
    \item \textbf{Overall Perception (OveP)}: A Likert scale value (1 to 5) denoting the overall sentiment of the text or article.
    \item \textbf{Country Perception (CouP)}: A Likert scale value (1 to 5) denoting the sentiment representing only the nationality or demonym in the text or article.
    \item \textbf{Diagnosis Parameter (DiaP)}: A binary answer to the question `Does the text snippet contain unreasonable, rude, or disrespectful content about the country in question?'
    \item \textbf{Toxic Parameter (ToxP)}: A binary answer to the question `Does the text snippet contain very hateful or aggressive content about the country in question?'
\end{itemize}

The metrics were developed by incorporating principles from sentiment, and opinion analysis \cite{liu2010sentiment, pang2008opinion}, as well as toxicity analysis \cite{borkan2019nuanced, do2019jigsaw} in natural language processing. The first two parameters, Overall Sentiment, and Country Perception seek to simulate the sentiment analysis performed by computational models to ``determine readers'' attitudes towards specific objects or entities' \cite{liu2010sentiment}. In contrast, the Diagnosis and Toxic parameters replicate the approach established by \citet{borkan2019nuanced} for identifying toxicity and hate speech in the text by asking human annotators to determine whether a given text contains unreasonable, rude, or disrespectful content, or very hateful or aggressive content, respectively. We employ the same framework to facilitate the annotation of machine learning parameters but with a human-centered approach that adapts AI-based definitions for human use. Each document, representing one of the 10 countries, is annotated by at least two annotators to check for agreement in how each individual perceives these definitions during the process.

\subsection{Interview Design}

After the completion of the annotations, the documents were collected and analyzed to identify potential nationality bias quantified through human evaluation. Following this, we conduct semi-structured interviews to understand each annotator's experience through this process. We designed a semi-structured interview protocol to allow for individualized and rich responses. 
%In general the questions centered around three areas: (i) annotation process interpretation (ii) impressions of the countries (both prior and developed) and (iii) AI generated texts and human written text. 

Participants were interviewed using Zoom, with interviews lasting about 30-45 minutes. As our goal was to collect answers that were individualized and open-ended, each participant spent as much time as they liked in answering each question, without interruptions. The interview was organized into several general sections:

\begin{enumerate}
    \item Grounding questions about the study and their perception of the annotation parameters.
    \item Deeper dive into their impressions about respective countries and if those impressions informed their annotation process.
    \item Revealing the sources of the text and studying how the particpants' perceptions changed.
\end{enumerate}

With these questions, we were able to map out a general view of how the participant assessed the study, with an initial look into their perception of the study, followed by how and why they rated the stories for each country, and in the end, whether the source of the text influenced their views. 

%As our goal was to collect individualized and open-ended answers, each participant spent as much time as they liked in answering each question without interruptions.

\begin{figure*}
\centering
\includegraphics[width=16cm, height=6cm]{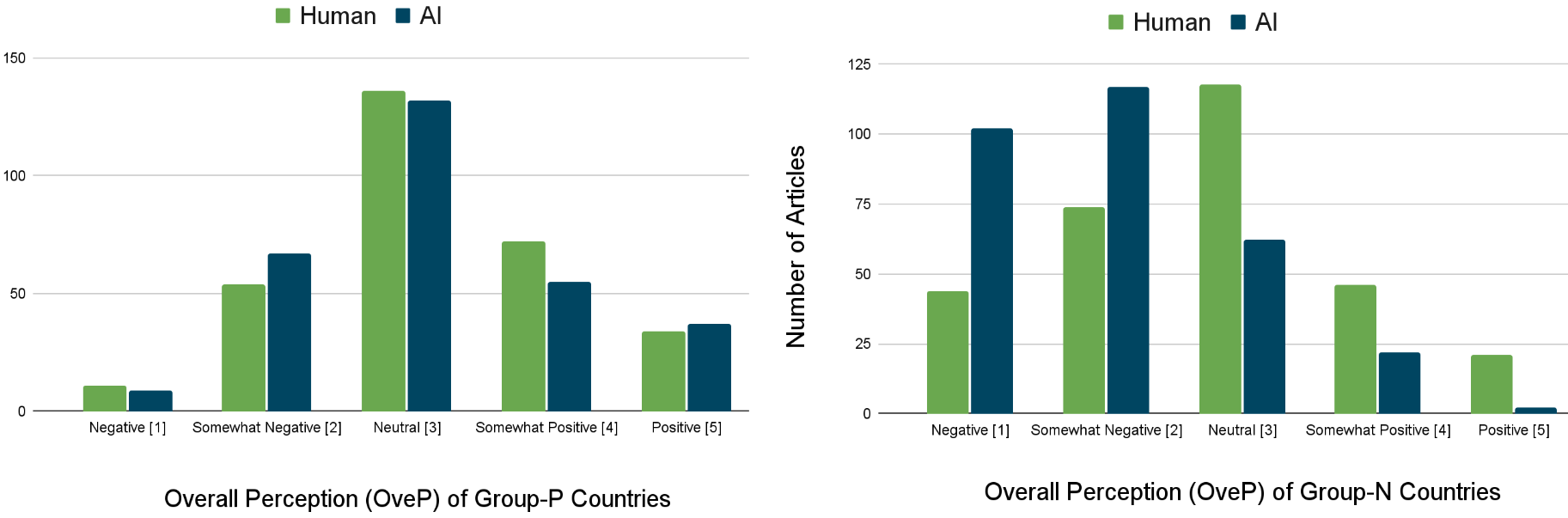}
\caption{Overall Perception (OveP) score of all articles grouped by the sentiment group of the countries.} \label{fig:OveP}
\end{figure*}

\section{Quantitative and Statistical Analysis of Annotation}

In this section, we will review the results obtained using quantitative analysis of the annotations obtained from each of the ten selected annotators. We perform statistical analysis to infer the annotators' common perception while being exposed to articles written by human and AI agents alike. 

\subsection{The Analysis of Sentiment}
The results presented in Table \ref{table: CouP} and Figure \ref{fig:OveP} highlight the difference in perceived sentiment between AI and human-written articles for Group-P and Group-N countries. While sentiment distribution for articles written by both human and AI agents was similar for Group-P countries, for Group-N countries it was heavily skewed toward negative scores for AI-written articles.  The mean score of \(\sim \)3 [Group-P: CouP[AI] = 3.28 , CouP[Human] = 3.17; Group-N: CouP[Hum] = 2.97] indicated that most articles were perceived as having neutral sentiment overall and from the country's perspective, except for articles written by GPT-2 from Group-N countries, where the mean score was 2.03 (somewhat negative). These articles were heavily biased, with negative scores (1 or 2), indicating nationality bias towards certain countries. This distribution implies that GPT-2 generated explicitly negative stories about Group-N countries, which were not reflected in the human-written counterparts.

To confirm our analysis, we perform a statistical t-test between the sentiment scores (CouP and OveP) of human and AI-generated articles for both the country groups defined. Our t-test revealed a highly significant difference in the scores annotated between AI and human articles in Group-N \textit{(CouP p-value = 4.87e-18, OveP p-value=2.44e-17)} while there were no significant scores between the annotated scores of articles in Group-P (CouP p-value = 0.2, OveP p-value=0.4). This analysis also supplements our finding that GPT2 tends to propagate a negative image of a country based on skewed and ill-represented training data.

\subsection{The Analysis of Toxicity}

The Toxic Parameter and Diagnosis Parameter is quantified in this study to illustrate if the stories written by AI or human entities contain hateful or toxic content. Our analysis of these parameters, presented in Table \ref{table: Toxic}, reveals that GPT-2-generated articles exhibit higher levels of the Diagnosis Parameter across both Group-P and Group-N. Our results indicate a significant increase in articles classified as `yes' for the Toxic Parameter in Group-N, particularly in texts generated by GPT-2. Our t-test revealed a significant difference in GPT-2 written and human written articles for the presence of only the Diagnosis Parameter in Group-P countries\textit{(DiaP p-value = 7.54e-18, ToxP p-value=0.25)} but showed high significant between human and GPT-2 written articles for the presence of both Diagnosis and Toxic parameter in Group-N countries \textit{(DiaP p-value = 4.89e-18, ToxP p-value=0.07)}. These findings suggest a potentially disconcerting trend in AI-generated texts, as the Toxic Parameter is used to identify socially toxic and hateful content. 

% Our results suggest that AI models like GPT-2 may exhibit such behavior more frequently for specific countries, highlighting the need for greater scrutiny of AI-generated texts.

\begin{table}[]
\begin{tabular}{|l|cc|cc|}
\hline
& \multicolumn{2}{c|}{\textbf{Group-N}}             & \multicolumn{2}{c|}{\textbf{Group-P}}             \\ \hline
& \multicolumn{1}{c|}{\textbf{Human}} & \textbf{  AI  } & \multicolumn{1}{c|}{\textbf{Human}} & \textbf{  AI  } \\ \hline
\textbf{DiaP} & \multicolumn{1}{c|}{32}             & \textbf{47} & \multicolumn{1}{c|}{11}             & \textbf{41} \\ \hline
\textbf{ToxC} & \multicolumn{1}{c|}{16}             & \textbf{23} & \multicolumn{1}{c|}{5}              & \textbf{9}  \\ \hline
\end{tabular}
\caption{Diagnosis Parameter (DiaP) and Toxic Parameter (ToxP) count of all articles annotated as `yes'.} \label{table: Toxic}
\end{table}

\subsection{Analysis of Adjectives}

%The analysis of adjectives used in stories can reveal the underlying themes and perspectives presented by the authors. 
In this section, we analyze the most common adjectives present in stories written by humans and the AI model, GPT-2, for countries in different groups, shown in Table \ref{table: adj}. The adjectives present in these articles are extracted using TextBlob \cite{loria2018textblob}. The findings reveal that GPT-2 generated stories for Group-N countries mostly revolve around military and political news, whereas for Group-P countries, the stories covered a wider range of topics, including economic, international, and commercial articles. Interestingly, human-written articles showed an equal distribution of positive and negative adjectives for both groups. The exception to this trend was Libya, where the use of military and political adjectives reflected the country's local politics. 
%The adjective analysis highlights the biased themes generated by GPT-2 for certain countries and the importance of analyzing language patterns in automated text generation systems.

\begin{table*}[]
\small
\begin{tabular}{|c|c|c|}
\hline
\textbf{Country} & \textbf{AI-written} & \textbf{Human-written} \\ \hline
Afghanistan & military, taliban, afghanistan, political, major & many, new, good, international, full \\ \hline
Finland & important, social, live, political, great & last, good, new, higher, less \\ \hline
France & good, new, great, different, understanding & new, last, independent, top, senior \\ \hline
Ireland & different, important, able, common, white, & last, new, first, best, personal \\ \hline
Libya & military, united, islamic, political, international & political, last, national, frozen, military \\ \hline
San Marino & different, good, cultural, political, civil & smallest, good, international, financial, social \\ \hline
Sierra Leone & military, political, civil, legal, humanitarian & new, young, commercial, special, national \\ \hline
Sudan & united, political, military, humanitarian, civil & economic, social, political, local, democratic \\ \hline
Tunisia & military, political, human, united. islamist & foreign, last, happy, former, diplomatic \\ \hline
United Kingdom & different, social, ethnic, cultural, conservative & private, national, short, financial, high \\ \hline
\end{tabular}
\caption{Top 5 adjectives for each country categorized by human and AI-generated model. } \label{table: adj}
\end{table*}

\subsection{Quantifying Nationality Bias}

Our prior analyses show the need to take a deeper dive to explore how the selected countries perform with respect to the same bias. To answer this, we quantify additional two metrics, \textit{Country Accentuation} (CA) and \textit{Overall Accentuation} (OA), as a measurement to help measure the impact of the bias generated by GPT-2. We formulate these parameters as follows:

\[Overall Acc [CA] = \sum\limits_{ove \in OveP}\left [{f(ove_{AI}) - f(ove_{Hum})}  \right ]\]
\[Country Acc [OA] = \sum\limits_{cou \in CouP}\left [{f(cou_{AI}) - f(cou_{Hum})}  \right ]\]

The metric \textit{Overall Accentuation} (\textit{OA}) measures the difference between how people perceive articles generated by GPT-2 for a selected group, $f(ove_{AI})$, and how they perceive articles written by humans for that same group, $f(ove_{Hum})$. The metric \textit{Country Accentuation} (\textit{CA}) is similar but measures the difference for a specific country. 
% Both metrics compare the average perception values of GPT-2 generated articles, $f(cou_{AI})$, and human-written articles, $f(cou{Hum})$. 
Table \ref{table: Bias} presents the results of the \textit{OA} and \textit{CA} metrics for ten countries. 
%The table displays the positive and negative accentuation exhibited by GPT-2 compared to human-written articles. 

%The findings indicate that for countries such as Sierra Leone and Tunisia, GPT-2 tends to convey a more negative image than what is reflected in the articles. Conversely, for countries such as Finland and San Marino, GPT-2 demonstrates a positive bias. Additionally, we observe a consistent pattern in the results for both the Country and Overall Accentuation parameters.

\begin{table}[]
\small
\begin{tabular}{|c|c|c|}
\hline
\textbf{Country} & \textbf{\begin{tabular}[c]{@{}l@{}}[CA] Country\\ Accentuation\end{tabular}} & \textbf{\begin{tabular}[c]{@{}l@{}}[OA] Overall\\ Accentuation\end{tabular}} \\ \hline
Sierra Leone     & -1.36                                                                   & -2.11                                                                   \\ \hline
Tunisia          & -1.13                                                                   & -1.10                                                                   \\ \hline
Sudan            & -0.77                                                                   & -0.61                                                                   \\ \hline
Libya            & -0.45                                                                   & -0.41                                                                   \\ \hline
United Kingdom   & -0.15                                                                   & -0.40                                                                   \\ \hline
France           & -0.08                                                                   & -0.04                                                                   \\ \hline
Ireland          & -0.05                                                                   & -0.05                                                                   \\ \hline
Afganistan       & +0.01                                                                   & -0.16                                                                   \\ \hline
San Marino       & +0.34                                                                   & +0.34                                                                   \\ \hline
Finland          & +0.49                                                                   & +0.20                                                                   \\ \hline
\end{tabular}
\caption{Country and Overall Accentuation value to show to calculate the bias in stories generated by GPT-2} \label{table: Bias}
\end{table}

\section{Qualitative Coding and Thematic Analysis}

The interviews of all the annotators were recorded and transcribed by the authors using a mix of automated software and manual checking. The transcribed interviews and textual data records were analyzed using \textit{analytic induction}, a mixture of deductive and inductive approaches \cite{robinson1951logical, znaniecki1934method}. While designing the interview, we knew that we are looking to understand if the annotators can identify nationality biases, prior experiences informing the annotation process as well as the annotation process impacting their impressions and annotator trust in AI-generated text. We did not disclose the intention of our study to the annotators and provided only details required to measure the `sentiment' and `toxicity' of the articles written by `unknown' sources. Our more detailed understanding of these issues emerged from our iterative review of the transcripts and is summarized in Table \ref{table: Quote} and below. 

\subsection{The Content - Group-P vs Group-N}

The annotation task revealed patterns in the type of writing that was observed by the participants. It was observed that the writing styles and themes of the texts were different for the Group-P and Group-N. We discuss below these differences and also the possible reasons for the same. These differences can have major implications in how the country is represented by the language generation models, and by extension, in the greater scheme of things. We initially saw this in our quantitative analysis of the adjectives, and in our interviews, we saw that the readers experienced the same difference in various ways.

\subsubsection{\textbf{Difference in writing themes}}

Some participants noticed significant differences in writing styles for Group-P and Group-N. The differences are highlighted in the examples below.  

%\textit{\say{They were quite difficult to comprehend structurally. but most of the themes that I noticed were about. You know Irish people and their sort of nationality, and how nationalism, I would say, but not in the negative sense, just their pride and nationalism towards their country, and how they are proud to be called Irish.} - P3}

\textit{\say{Finland which talked about them being proud people, immigration, and that was something I was noting. But for Sudan maybe it evoked a little bit of sadness maybe, pity.} - P5}

\textit{\say{They were not in line with the knowledge that I had, because they talked a lot about Finnish pride, immigration stuff, and also like just about their culture and stuff} - P4}

\textit{\say{A lot of the Libya articles in particular are focused on the Civil War, and events following that, and focused on the violence that took place. Whereas the UK. I mean, there were a few that focused on the troubles and shared some similarities and discussed kinds of terrorist tax. But to a much lesser extent. Yeah, a lot of the UK articles focus on kind of political changes. The Libyan articles, even when they kind of focus on the country's government, it's much more on the various military aspects of it where the various groups controlled the country.} - P1}

As observed by P3, the scales were relative for both countries. This was due to the nature of topics covered in the articles for the individual countries. The participant saw benign themes for Ireland (Group P), whereas there were stronger themes such as \textit{terrorism} when it came to Tunisia (Group N). Further P3, P4, and P5, saw that the topics referred to topics like \textit{proud people} and \textit{good immigration system}. Following this, the participant further reflected on how annotating within a document led to further removal of objectivity in the ratings. 

\textit{\say{So, going back to Ireland from Tunisia, I would sort of find myself rating things probably a little higher than I would have rated them without having Tunisia in contex, because I feel like the things I read for the Tunisia ones were so negative that in comparison the things happening in the Ireland article weren't as negative. So I would find myself, you know, rating them 1. I would have otherwise maybe rated it 2.} - P3}

\textit{\say{I feel like the 1 to 5 <annotation scale> in Ireland, and the 1 to 5 in Tunisia were on slightly different scales, because the Ireland articles which were coded as like 2 or 1 <by me> were talking about relatively easier topics as opposed to the Tunisia articles were talking a lot about terrorism and attacks and violence, and a bunch of those things, and even like coups and such.} - P3}

Here we observe two interesting findings. The first is the fact that all the annotators found a significant difference in themes between the articles present in Group-P and Group-N. As per our adjective analysis, the articles that had differences in theme were the ones that are generated by GPT-2 for the Group-N countries. This shows that these negative articles had more impact on the readers as compared to the rest. We will discuss this in detail in the following section. The second is the fact that the annotators perceived and used their implicit learning in annotating a text concerning sentiment and toxicity. So, to conclude we saw that participants were perceptive of the nationality biases. We will discuss in aspect further in the discussion section.

\begin{table*}[]
\begin{tabular}{|c|c|c|c|}
\hline
\textbf{Tag}                                                                        & \textbf{Theme}                                                              & \textbf{Quote}                                                                                                                                                                                                                                                                                                                                                                                                & \textbf{\begin{tabular}[c]{@{}c@{}}Annotator\\ ID\end{tabular}} \\ \hline
\begin{tabular}[c]{@{}c@{}}Difference in \\ writing themes\end{tabular}             & \begin{tabular}[c]{@{}c@{}}The Content - \\ Group-P vs Group-N\end{tabular} & \begin{tabular}[c]{@{}c@{}}`Most of the themes that I  noticed were about\\ Irish people and their sort of nationalism,\\ I would say, but not in the negative sense, just \\ their pride and nationalism towards their country, \\ and how they are proud to be called Irish.'\end{tabular}                                                                                                                  & P2                                                              \\ \hline
\begin{tabular}[c]{@{}c@{}}Prior Opinion \\ Clouding Annotations\end{tabular}       & \begin{tabular}[c]{@{}c@{}}The Content - \\ Group-P vs Group-N\end{tabular} & \begin{tabular}[c]{@{}c@{}}`So whenever I was marking for Afghanistan, \\ I was being extremely careful, because I'm like, \\ I don't want my impression or my understanding\\  of the country to be in the way of coding.'\end{tabular}                                                                                                                                                                      & P8                                                              \\ \hline
\begin{tabular}[c]{@{}c@{}}N-Countries elicit\\ more emotions\end{tabular}          & \begin{tabular}[c]{@{}c@{}}N-Countries elicit \\ more emotions\end{tabular} & \begin{tabular}[c]{@{}c@{}}`And essentially, even though they have \\ problems with England, it's a fairly developed \\ country that has fairly well-structured systems, \\ there was one text about this accident that kill 3 \\ people is so rare. But still it's such a big deal which \\ you know, just reinforces the kind of text you would \\ expect from a relatively developed country'\end{tabular} & P3                                                              \\ \hline
\begin{tabular}[c]{@{}c@{}}Distrustful of the\\ Text Presented\end{tabular}         & \begin{tabular}[c]{@{}c@{}}Influence of \\ Text Sources\end{tabular}        & \begin{tabular}[c]{@{}c@{}}`I wasn't getting any impression from those texts \\ because I didn't know where they are coming\\  from or who have written them.'\end{tabular}                                                                                                                                                                                                                                   & P9                                                              \\ \hline
\begin{tabular}[c]{@{}c@{}}Poorly written AI\\ generated text\end{tabular}          & \begin{tabular}[c]{@{}c@{}}Influence of \\ Text Sources\end{tabular}        & \begin{tabular}[c]{@{}c@{}}`Just the way it's written like it starts in one place\\  and ends somewhere else. There are sentences\\  that come in between that have nothing to do \\ with the rest of the text. It doesn't feel like it's\\  going anywhere in particular. It's going in \\ like 5 different directions.'\end{tabular}                                                                        & P8                                                              \\ \hline
\begin{tabular}[c]{@{}c@{}}Diagnosis Parameter\\ vs Toxicity Parameter\end{tabular} & Study Perception                                                            & \begin{tabular}[c]{@{}c@{}}`So anything that wasn't hateful or directed. \\ It sort of goes to the intensity of what the text \\ is saying. I think if it was very intense, then\\ it sort of appeared toxic to me'\end{tabular}                                                                                                                                                                              & P6                                                              \\ \hline
\end{tabular}
\caption{Themes obtained during the thematic analysis along with their respective additional quotes.} \label{table: Quote}
\end{table*}

\subsubsection{\textbf{Prior Opinion Clouding Annotations}}

Another aspect that we wanted to understand if prior opinions played a role in their annotation process. We saw that some participants reflected on how their prior opinion affected the way they annotated the texts. While some participants consciously try to mitigate the impact of their personal thoughts on the texts, others were aware of the bias and realized that it might have swayed how they rated a particular country. 

%\textit{\say{So whenever I was marking for Afghanistan, I was being extremely careful, because I'm like, I don't want my impression or my understanding of the country to be in the way of coding.} - P8}

\textit{\say{Unfortunately, as a <redacted>, I did have prior held beliefs about Afghanistan, and they weren't overtly negative. I just understand Afghanistan as a country where the US has had some unfortunate dealings and as such and that's unfortunately my only familiarity with the country is things that I experienced here in the <redacted> media and so my perception of them isn't overly negative. However, it is skewed by what has essentially been peddled down <redacted> citizens’ throats.} - P10}

Additionally, we also saw that the beliefs that people held for N Group countries were reinstated, or in the case where they did not have one, they developed one based on the articles they were exposed to during the annotation process. 

\textit{\say{I <am> familiar.. starting with the Civil War following the Arab spring in 2010, and kind of with the fall of Gaddafi and the various pieces of government <Libya> that came together afterwards. And kind of my impression would be that, you know there's a fairly fragile government in place, but still not a strong state that's still struggling in many ways.} - P1}

\textit{\say{I think it was sort of a contrast from Finland to Sudan. I for Finland, I thought it was a wealthy country, but for Sudan I thought it was probably a poor country with not a lot of resources.} - P5}

When we asked the participant if this impression was upheld, the response is:

\textit{\say{No, I'd say it was largely in line with my previous prior impression.} - P1}

We saw a similar trend when it came to P-Group countries. The participant had a general impression of the country that was upheld during the process of annotation as well. Another thing to note here would be that most of these quotes revolve around the impression that is being carried for the N-Group countries. 

% \textbf{<there are more quotes and points to make under this if space permits>}

\subsubsection{\textbf{News like for Group-N vs Opinion pieces for Group-P}}

We saw that most of the participants felt like the language of articles that spoke about N-Group was more like a news article or report. They talked about major happenings in the area concerning war and terrorism. While when it came to P-Group, there were story-like articles that detailed just thoughts and opinions of people about the country. 

%\textit{\say{I saw something like news, a piece of news article} - P9}

\textit{\say{You know it's kind of from the tone. The writing style of it most generally seems perfectly fine to be a piece of news} - P1}

\textit{\say{Tunisia...it's a little more difficult because all of them are written in the style of news article. But at the same time, like there are some which you know does give like more of an opinion, or feels like there’s more of judgment and sentiment in the GPT-2 generated articles.} - P3}

\textit{\say{Finland which talked about them being proud people, immigration, and that that was just I, I was noting maybe similarities in that with the other countries.} - P5}

\textit{\say{I thought there was a lot of deep culture associated with San Marino, and I I thought it was overall pretty positive about the culture of San Marino.} - P6}

The nature of articles also points to the tone and sentiment that people might perceive of the content. It is important to understand the implied effect this might have on the impression of the country for the annotators going forward. 

\subsection{\textbf{N-Countries elicit more emotions}}

We saw that people exhibited more strong emotions when it came to talking about N-countries. Participants were more `moved' and `impacted' by the content they saw for N-countries than P-counties. 

\textit{\say{I think the Sierra Leone articles were more moving for me, that I felt more strongly about as compared to the UK. I would have a harder time sympathizing with that country than Sierra Leone.} - P7}

\textit{\say{It said that the increasing population of Islamic countries are becoming a problem and that…that personally offended me I think. Yeah for me, I think that was inciting. Yeah, that was an emotion, I think. Other than that, San Marino felt like some country I would like to visit.} - P8}

\textit{\say{Through the passages it hit me how bad or at least I don't know if how recent all of these passages are. But yeah, it made me think more deeply about really how bad the situation seems to be there.} - P7}

The feelings participants shared about the news articles discussed for the P-Group differed largely from the ones discussed for the N-Group. For example, we can find below that the participant has an impression and expectations of UK (P-Group). These expectations are consistent with their belief such that even the negative news does not have a significant emotional impact on them.  

%\textit{\say{It’s a Western European country, is, in a union with England. And essentially, even though they have problems with England, it's a fairly developed country that has fairly well-structured systems, and probably do you know, like, there was one text about like, oh, like this accident that kill 3 people is so rare. But still it's such a big deal which you know, just reinforces the kind of text you would expect from a relatively developed country.} - P3}

Additionally, one of the participants observed that rating the countries one after another interspersed affected the rating. This was an especially interesting observation considering annotators are often presented with randomized organization of content. 

\textit{\say{I would feel worse about what was happening in Tunisia after reading the things from Ireland. So I don't know if that would influence me, really, but I feel like it made me feel it evoked more emotions in me than it did after say I had read 20 Tunisia articles continuously.} - P2}

Additionally, we say that major emotion was attached to N-Group countries and people's perception of P-Group countries did not elicit much emotion. This is to say that the content for the P-Group country was neutral enough to not tap into the emotional side of the participants. We see the same with the quantitative analysis of the texts as well.

\subsection{\textbf{Influence of Text Sources}}

The text to be annotated was shared with the participants as randomized text containing both AI-generated and human-written text. The participants were made aware of the exact text sources for each entry during the interview. The participants were given some time to reflect and consider the prompts. When the sheets describing the text sources were presented to the participants, we studied their reactions to the same. We discuss below these responses under three different headings:
% \textbf{<add further description based on the themes below>}

\subsubsection{\textbf{Distrustful of the Text Presented}}

Some of the participants who annotated were apprehensive about the content of the text given that they were unaware of the source of the text. They had this impression even before they gained insights into the source of the text.  

\textit{\say{I wasn't able to judge if something was unreasonable because I didn't have access to ground truth about those passages.} - P7}

%\textit{\say{I wasn't getting any impression from those texts because I didn't know where they are coming from or who have written them.} - P9}

With these participants, we see how they would not rely on an unverified source unless they do background checks. In contrast to this, we observed that some did believe what they read and did not question it.

\textit{\say{I would never think <of the source>. I would probably just believe it. So it seemed like something possible, something that could happen.} - P8}

\subsubsection{\textbf{Poorly written AI generated text}}

When the participants were asked upon reflection if they felt any of the text was AI-generated, we saw that they responded that in most cases reflecting back, they could notice the difference. We discuss below the elements that helped them identify this difference.

\textit{\say{They weren't free-flowing text and it also stood out to me as not a well-written article.} - P7}

\textit{ \say{There were a lot of grammar mistakes, and sometimes the sentences didn't structurally make sense, or one sentence wasn't related to the next one.} - P2}

A common theme was that the grammar from AI written text suffered heavily. Additionally, participants strongly associated the word `nonsense' with most of the text that was generated by AI. 

\textit{\say{There were a lot of grammar mistakes, and sometimes the sentences didn't structurally make sense, or you know, one sentence wasn't related to the next one.} - P3}

%\textit{\say{Just the way it's written like it starts in one place and ends somewhere else. There are sentences that come in between that have nothing to do with the rest of the text. It doesn't feel like it's going anywhere in particular. It's going in like 5 different directions.} - P8}

\subsection{Study Perception}

Human annotators are widely used in AI systems, especially in the case of human-evaluated systems. We used this section of the interview to highlight the happenings inside an annotator's and a reader's mind from a different perspective even when detailed instructions are presented to them. We observed the following three themes.

\subsubsection{\textbf{Individualistic Perception of the Goals}}

When presented with a set of information, the participants each perceived the aim of the study in a different way. The perceptions ranged from `nuanced sentiment analysis' to `understanding if something needs to be taken off the internet'. While the perceived goals were not significantly off, not capturing the exact goal can lead to inaccurate annotations. The following quotes help us understand the differences in the perceived goals.

\textit{\say{You might be looking for detection of hate speech for different countries.} - P10}

This participant felt that we are trying to identify \textbf{`hate speech'} in the articles given. While there was a focus on hate speech detection as well, the participant may not have focused on the more nuanced differences when it came to bias identification and hate speech detection. The key is that while bias can be observed through hate speech, it is not the only way it can. 

\textit{\say{To me, it seemed like what you were looking for was how sentiment sort of mixed in, or what are the correlation between sentiment and the other parameters, and that I had to fill out for each country.} - P5}

Similarly, in the above quote from the participant, we see that the focus is only on sentiment capturing. 

\textit{\say{Yeah, I guess it seemed like just trying to assess the perception of 2 different countries in a set of scraped news articles.} - P1}

So, we can see that on the surface, they appear to be similar objectives; however, to a trained researcher's eye, they may as well be three different research projects. For our research, we intend to capture both the perception of human readers to text generated by AI and human sources and understand how machine learning concepts translate to human studies. Through this, we see that it is not intuitive to use machine learning fundamentals without additional  aspects to make it human-study-friendly.

\subsubsection{\textbf{Overall vs Country Perception}}

As described before, we used two parameters : (i) Overall Perception and (ii) Country Perception. We asked the participants to identify the difference between the two parameters.

\textit{\say{<Overall sentiment> takes into just the entire tone of the paragraph where the country perception is specifically the tone towards that perceived country.}-P6}

\textit{\say{I kind of thought like country perception is like, I’m trying to look for what people think of a country as opposed to like the content of what happens in a certain country you are like} - P3}

Participants indicated similar perceptions of the definitions. They were mostly in line with the ones given to them. 

\textit{\say{Sometimes it comes off as being neutral or positive, but there must have been something negative about the country mentioned, so that was distinguishing that I made <between overall and country perception>. The article itself might have been wrapped up very positively.} - P8}

\subsubsection{\textbf{Diagnosis Parameter vs Toxicity Parameter}}

While the definitions of both parameters used very different adjectives to define them, the participants reduced them to higher and lower thresholds. This was the intention of the study to understand the different degrees of toxicity.  

\textit{\say{But I think what I did was for diagnosis. It was like something little, not okay. For toxic, it was like, okay, this is problematic.} - P8}

%\textit{\say{an intuitive meaning, so anything that wasn't hateful or directed. I, I think it. It sort of goes to the intensity of, what the text is saying. I think if it was very intense, then it sort of appeared toxic to me} - P6}

\section{Discussion}

% Our work tries to understand the presence and interpretation of nationality bias in LLM-generated text when contrasted with Human written text. In this section, we will discuss and interpret the results obtained from our quantitative and qualitative analysis to answer the questions we presented initially.

% Past work has indicated that the quality of annotations is often assessed as the measure of interannotator agreement \cite{campbell2011introduction}. The intuitive idea of quality assurance through majority agreement via humans providing the same annotation for the same examples. Prior work has indicated the pitfalls in the current human annotation process \cite{aroyo2015truth}. Through this work we took a deeper dive into how bias plays out when we use human annotators to resolve and mediate mislabeled entries of training data. 

% The two fold approach allowed for us to investigate the "what they say" along with the "what they do". The results described above are indicative of the possible explanations of the annotation scores assigned. Specifically in the case of bias mitigation, it is essential to unwrap the process that annotators undergo while annotating. 

% Through our findings we see... \textbf{(link the qualitative and quantitative findings)}

\subsection{Quantified Human Perception of Bias}

Our results show that the annotators were able to identify the negativity and toxicity in the GPT-2 generated texts, even without knowing the source. Our Country and Overall Accentuation metrics show that the GPT-2 generated texts for the countries from Group-N showed a significant difference from the rest of the articles. They were perceived to be more negative and toxic than their human-written counterparts. Our adjective analysis also shows that GPT-2 has a perception of the Group-N countries that do not agree with its human-written counterpart, which shows an equal representation of positive and negative countries. GPT-2 written texts for Group-N countries heavily exaggerate military and war-like themes. We observed that our annotators could recall sharper details about these themes than the positive ones. This is indicative that negative texts have a more substantial impact on human memory. It also relates to the notion of implicit memory, which previous studies have indicated that individuals tend to prioritize the recollection of negative stereotypes over positive ones \cite{perdue1990evidence, hense1995implicit, banaji2016blindspot}, resulting in an implicit inclination toward negative bias in memory retention. Consequently, our findings show the societal consequences of GPT-2 amplifying negative biases, as the false negative bias can lead readers to capture erroneous information.

% Representational harm is defined as the harm that arises when a system represents some social groups in a less favorable light than others, demeans them, or fails to recognize their existence altogether, and allotted harm is defined as the harm that arises when a system allocates resources or opportunities unfairly to a social group \cite{blodgett2020language}. 

Our work shows that the biases depicted by models like GPT-2 may have social impacts that also translate to representational and allotted harm. These results show that if such text generation models are used in a sociotechnical system, the biases identified can also be translated to potential harm. From the framework of harm postulated by \citet{dev2021onmeasures}, we can see how such behaviors can lead to harmful social behaviors such as \textit{stereotyping, disparagement} and \textit{erasure} where certain nationalities are oversimplified, evaluated as `lesser' and underrepresented by the model respectively. 

\subsection{The Impressions of the Texts}

Our qualitative results add explainability to our findings in the quantitative section. We use our qualitative results to dive deeper into `why' the annotators answered the way they did. The results of our qualitative analysis and interview sessions reveal that the biases identified by the annotators did indeed create an impression of the country they were annotating, which in turn influenced their annotations moving forward. This highlights the instant impact of biases and their potential to shape how we view the world. This underscores the immediate influence of biases and their capacity to shape our perception of the world. 

Interestingly, none of the annotators could explicitly identify the skewed perception of the country or that they were reading articles written by AI models until prompted to do so. Our analysis indicates that annotators were only able to implicitly identify bias by measuring the text as negative or toxic. This phenomenon can pose an issue as the biases in AI-generated content can remain unnoticed and continue to influence a reader's perceptions.

Another critical finding of our analysis was the phenomenon of `AI hallucination', where AI models provide confident responses that seem faithful but are nonsensical in light of common knowledge \cite{alkaissi2023artificial}. A number of participants (prior to being prompted about the text source) mentioned that they felt that some of the passages were hard to follow and did not make any sense to them. They often reported that while the text began talking about one topic, the next topic would not be in line. Our study indicates that text generated by AI models tends to be influenced by AI hallucination, leading to a more radical and opinionated tone. This behavior makes AI-generated content more likely to mislead, as it is written in a confident and authoritative tone that can be perceived as factual by the reader \cite{ng2023batch, alkaissi2023artificial}. 

%This behavior was not identifiable by the annotators until scrutinized showcasing that this behavior had an effect on their perception of the country.

%We also highlight the issue of trust in AI systems. Some annotators expressed confidence in the system's ability to be trained on `all data on the internet', and this perception is problematic as it assumes that the training dataset used is representative of all. In reality, the dataset used to train AI models is highly dependent on the data created by internet users, which may not be representative of the world. 
%As such, it is essential to recognize that AI models are not free from biases, and the data used to train them can shape their outputs significantly. 
We also notice that the language style used by AI models correlates with how the model views a country. Specifically, GPT-2 generated stories about Group-N countries in a manner reminiscent of news articles, while it tends to present Group-P countries with a tone resembling opinion texts.
% Specifically, we found that the GPT-2 writes stories for Group-N like news articles and for Group-P like opinion texts. 
% Specifically, we found that GPT-2 uses two varying writing styles: news and opinion texts, to describe countries from Group-N and Group-P. 
% This finding is critical to our understanding of how the model misrepresents information. 
This finding helps us understand how the model perceives the country and the associated information.
An opinion piece by it's very nature is perceived by the public as someone's opinion and not the ground truth. However, when N-Group country information is represented as news article it appears to be the ground truth. GPT-2 generated text further used terms like `the BBC' to validate the idea being conveyed. 
This further leads to propagation of the AI authority phenomenon. This can have far reaching impacts when used in social scenarios. This finding underscores the importance of considering language style and its potential impact on perception when working with AI-generated content.

\subsection{Human Perceptions of Automatic Indicators}

Our qualitative analysis highlights the need for an interdisciplinary approach to bias identification in AI and NLP models. While previous studies have primarily focused on using automated evaluation to measure and quantify bias \cite{bolukbasi2016man, kiritchenko2018examining}, our paper presents a unique perspective by using human annotators to identify and attempt to quantify bias. In the process of conducting the interviews, we found that every participant had a different and unique perception of the goals of the study and the metrics they were asked to calculate. Although we use automatic indicators, our findings reveal that humans have differing perceptions on the given definitions of these parameters. These perceptions that do not always match those provided in the field of AI. Our readers, who could identify and rate bias in sentiment, viewed sentiment and differently, as seen by the low Cohen-Kappa values (OveP = 0.34, DiaP = 0.38). Similar values were observed for toxicity as well (DiaP = 0.18, ToxP = 0.38). These results show that it is necessary to consider differing human interpretations when defining and understanding computation-based parameters that can have a direct impact on human perceptions.

\section{Conclusion}

The paper uses human evaluation to explore nationality bias in a text generation model (GPT-2). The research uses the NLP sentiment and toxicity framework through human annotators to quantitatively analyze the presence of nationality bias. The findings reveal that the text generation model accentuates negative bias towards certain countries while demonstrating positive bias toward 'well-represented' countries. Using interviews, the study investigates how readers interpret articles generated by GPT-2. The interviews show that negative stories generated about certain countries had the most emotional impact on readers. However, some readers found such articles informative, informing them of a new aspect of the country. The study also found that participants were more welcoming of such technology after the disclosure that the articles were generated by both human and AI agents, as they were intended to 'mimic human behavior and biases.' The paper highlights the harmful impact of such technology if not used appropriately. It can enhance a country's skewed perception while maintaining the majority's viewpoint, leading to misinformation and stereotyping. 
% Overall, our paper contributes to the growing body of research on identifying bias in text generation models and emphasizes the importance of considering the human perspective in developing and using these models.

\bibliographystyle{ACM-Reference-Format}
\bibliography{sample-base}


\end{document}